\documentclass[11pt]{article}

\usepackage[preprint]{acl}

\usepackage{times}
\usepackage{latexsym}
\usepackage{amsmath,amssymb}

\usepackage[T1]{fontenc}
\usepackage[utf8]{inputenc}

\usepackage{microtype}

\usepackage{inconsolata}

\usepackage{graphicx}

\title{KwaiSQL: Preference-Optimized Selection for Enhanced Text-to-SQL Generation}

\author{
  Danyang Zhao\textsuperscript{\dag},
  Wenbo Zhang\textsuperscript{\dag},
  Yan Li,
  Xiaohui Li,
  Fei Chen, 
  Song Gao,
  Penghui Chen\textsuperscript{*},
  Ruining Wang\textsuperscript{*} \\
  Kuaishou Technology \\
  \texttt{\{zhaodanyang06,zhangwenbo09,liyan15,lixiaohui03\}@kuaishou.com} \\
  \texttt{\{chenfei08,gaosong,erdong,wangruining\}@kuaishou.com} \\
  \textsuperscript{\dag}Equal contribution.
  \quad
  \textsuperscript{*}Corresponding authors.
}

\begin{document}
\maketitle
\begin{abstract}
Text-to-SQL systems increasingly combine candidate generation with
query selection. When these two roles share a model, can learning
to select better queries also improve generation? Both tasks require
judging how faithfully SQL operations express a natural-language
question, making candidate comparison a potential source of
supervision for generation. We investigate this connection by
training a shared model for SQL generation and selection. Starting
from a tool-using generator trained with execution rewards, we
apply preference learning to candidate selection and use the
resulting checkpoint for both roles. We analyze how selection
updates can benefit generation through shared parameters and
evaluate the generated candidate pools before selection, using
the same tools and sampling settings. On BIRD Dev, selection
preference training raises generation-only pass@16 from 81.79\%
to 82.83\%, increasing the coverage of questions with at least one
correct candidate. The resulting model achieves 74.67\% execution
accuracy when used for both generation and selection. These
findings show that learning to compare SQL candidates provides
a training signal that improves candidate generation and raises
the ceiling for final query selection.Our model is publicly available at
\url{https://huggingface.co/Kwai-AutoSQL/Kwai-AutoSQL-27B}.
\end{abstract}

\section{Introduction}
\label{sec:introduction}

Text-to-SQL enables users to access relational databases through
natural language. Answering a question requires a model to connect
user intent with database schemas and contents, and express that
interpretation as an executable query. Spider emphasizes
generalization across database schemas \citep{spider}, while BIRD
highlights the importance of grounding queries in database values
and domain knowledge \citep{bird}. These demands make
semantic judgment central to SQL generation: the model must
determine which database operations faithfully represent the
question.

Recent approaches improve SQL predictions through both generation
and selection. Task decomposition and contextual grounding
support query construction and refinement \citep{din-sql,chess}.
Execution-based reinforcement learning further strengthens generation by rewarding queries that return correct results, as
demonstrated by Arctic-Text2SQL-R1 \citep{arctic-text2sql-r1}.
Candidate selection provides another route: MCS-SQL generates
queries using multiple prompts and chooses among them through
multiple-choice selection \citep{mcs-sql}. Learned selectors assess
alternative queries through pairwise comparison or joint critiquing
\citep{chase-sql,msc-sql}.
Together, these approaches establish generation and selection as
two closely connected components of Text-to-SQL systems.

Research on self-rewarding language models connects generation
and evaluation within a single model, improving both instruction
following and reward assessment through iterative training
\citep{self-rewarding}. In Text-to-SQL, this connection extends to
the semantic decisions underlying the two tasks. Generating a query requires deciding how to represent the
question through schema elements and SQL operations. Selecting a
query requires assessing those same decisions across alternative
solutions. Candidate comparison makes these alternatives explicit,
providing a setting in which the model can learn to distinguish
faithful interpretations from plausible errors. This relationship
motivates our central question: \emph{can learning to select better
SQL queries also improve the model's ability to generate them?}

\begin{figure*}[t]
    \centering
    \includegraphics[width=1.0\linewidth]{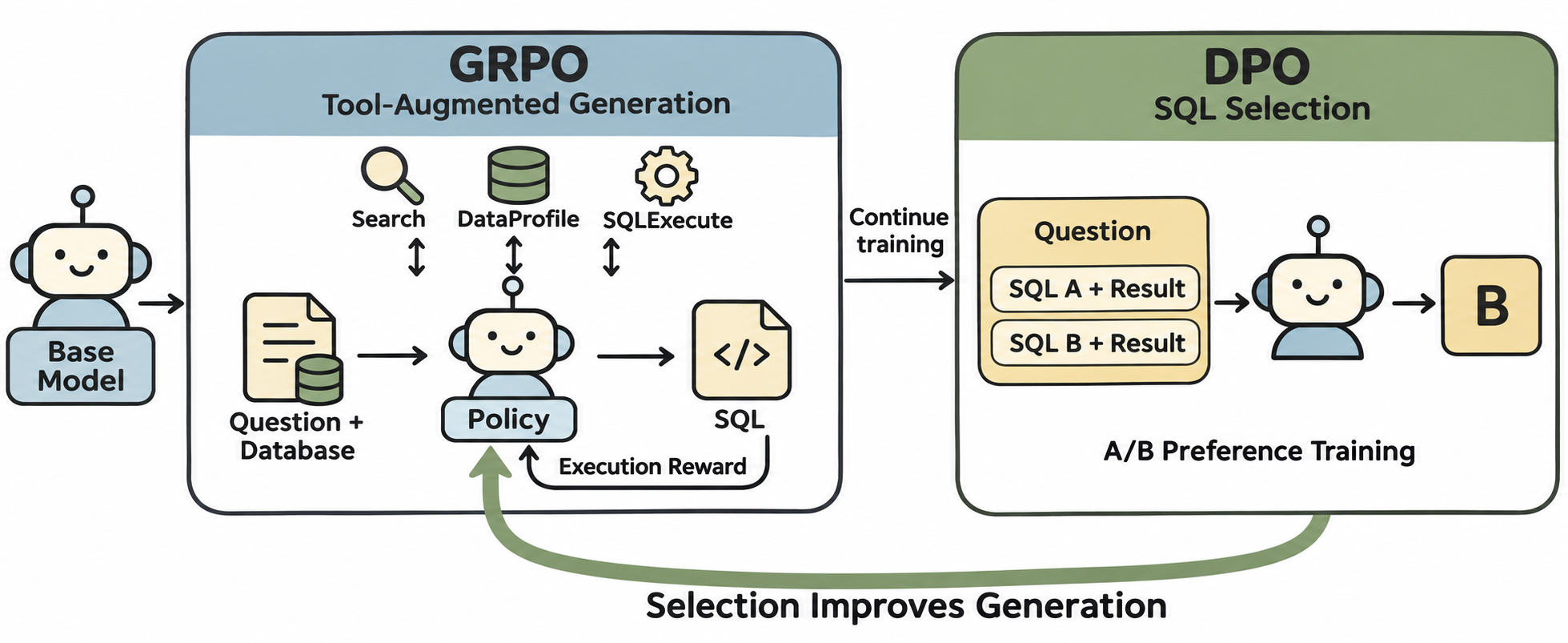}
    \caption{Learning to select SQL queries improves query generation.
    A tool-using generator is first trained with execution rewards,
    then optimized to choose between candidate queries through
    preference learning. The resulting checkpoint serves both
    generation and selection; its generation-only pass@16 measures
    correct-candidate coverage before selection.}
    \label{fig:main}
\end{figure*}

We investigate this question by connecting SQL generation and
selection through a shared model (Figure~\ref{fig:main}). Starting
from a generator trained with database tools and execution rewards,
we optimize the model to choose between candidate queries using
preference learning \citep{dpo}. We then use the resulting checkpoint for both
generation and selection. This design allows judgments learned
through candidate comparison to influence subsequent query
generation through shared parameters. Our analysis characterizes
this transfer in terms of the alignment between selection updates
and directions that increase the probability of generating correct
queries.

Experiments on BIRD show that selection training improves the
coverage of correct queries during generation. After preference
training, generation-only pass@16 increases from 81.79\% to 82.83\%,
measured before candidate selection. The resulting model achieves
74.67\% execution accuracy when used for both generation and
selection. These findings demonstrate that learning to choose can
improve the candidate pool itself, raising the ceiling for the
final prediction and connecting selection training to improved
SQL generation.

Our contributions are threefold:
\begin{itemize}
    \item We investigate transfer from SQL selection training to
    query generation, evaluating its effect on candidate coverage
    before selection.
    \item We develop a shared-model approach that connects
    execution-reward training with selection preference learning
    and uses the resulting checkpoint for both roles.
    \item We provide empirical evidence on BIRD and a
    shared-parameter analysis of how selection updates can
    benefit generation.
\end{itemize}

\section{Related Work}
\label{sec:related-work}

\paragraph{Agentic Reinforcement Learning for Text-to-SQL.}
SQL execution provides a verifiable reward signal for training.
Arctic-Text2SQL-R1 optimizes SQL generation using the execution
correctness of the final query \citep{arctic-text2sql-r1}.
Recent work extends reinforcement learning to multi-turn database
interaction, training models to retrieve information, execute queries,
and adapt their subsequent actions to feedback.
MARS-SQL trains a generation policy in a ReAct-style loop
\citep{mars-sql}, SQL-Trail adapts the interaction budget to question
difficulty \citep{sql-trail}, and FineStep derives training signals
for intermediate actions from structured tool outputs \citep{finestep}.
Following this line of work, we equip our model with schema search,
data profiling, and SQL execution tools, and train a multi-turn
generation policy with Group Relative Policy Optimization (GRPO) using execution rewards.

\paragraph{SQL Candidate Selection.}
Given multiple SQL candidates, one approach is to vote based on their
execution results. EGV-SQL groups candidates that return the same
result and selects the answer with the most support \citep{egv-sql}.
Model-based selectors judge which candidate best answers the question.
CHASE-SQL uses a separate model for pairwise selection, for example
generating queries with Gemini 1.5 Pro and comparing them with a
fine-tuned Gemini 1.5 Flash \citep{chase-sql}.
Generation and selection can also share a backbone while using
separately trained checkpoints. In its Mistral-7B configuration,
MSc-SQL uses one checkpoint to generate candidates and another to
assess them using their execution results \citep{msc-sql}.
MCS-SQL uses the same GPT-4 model for both stages, changing the
prompt to switch from query generation to multiple-choice selection
\citep{mcs-sql}.

\paragraph{Generation and Evaluation in Shared Models.}
Learning to evaluate candidate solutions is also a route to
improving reasoning systems. In mathematical reasoning, trained
verifiers rank sampled solutions to determine the final answer
\citep{training-verifiers}. Self-Rewarding Language Models bring
generation and evaluation into a single model: the model judges
its own responses to construct preferences for iterative training,
improving both instruction following and reward assessment
\citep{self-rewarding}. These results motivate examining how the
two capabilities interact during training. We study this interaction
in Text-to-SQL by fine-tuning a generator on candidate-selection
preferences and then evaluating its generated queries before
selection. This evaluation measures how learning to choose changes
the candidate pool produced by the shared model.

\section{Method}
\label{sec:method}

We first train a tool-using SQL generator with execution rewards,
then optimize its preferences over pairs of SQL candidates.
The resulting checkpoint performs both generation and selection.
Our central question is how learning to select changes the
model's ability to generate correct queries.

\subsection{Tool-Augmented SQL Generation}
\label{sec:sql-generation}

We initialize our policy from Qwen3.8-27B. For each database,
we generate natural-language descriptions of tables and columns
and supplement them with example values. The policy interacts
with the database through three tools:
\texttt{search} retrieves relevant schema information,
\texttt{dataprofile} provides data profiles, and
\texttt{sqlexecute} executes SQL and returns feedback.
These tools allow the model to gather information and refine
queries over multiple interaction steps.

We sample 16 trajectories per training question and retain
questions with both correct and incorrect outcomes.
An external teacher reviews the trajectories and filters out
examples with suspected reference SQL errors or ambiguous
targets. We then optimize the policy with GRPO \citep{grpo},
using final-query execution correctness as the reward.
This checkpoint initializes the subsequent preference training.

\subsection{Learning to Select SQL}
\label{sec:sql-selection}

\paragraph{Selection preferences.}
We sample SQL queries on the training split and construct
pairs of distinct candidates for the same question.
Each selection prompt contains the question, two candidates
labeled \texttt{A} and \texttt{B}, and summaries of their
execution results. The preferred response identifies the
better candidate, and the rejected response identifies the
other candidate.

Presenting both candidates gives the model a direct comparison
between alternative solutions. For example, two executable
queries may differ in whether they count rows or distinct
entities. Choosing between them requires relating the
question to the aggregation expressed by each query.

\paragraph{Preference optimization.}
Let $x$ denote a selection prompt and $y^+$ and $y^-$ denote
the preferred and rejected answer labels. We define the
selection margin as
\begin{equation}
s_\theta(x)
=
\log \pi_\theta(y^+ \mid x)
-
\log \pi_\theta(y^- \mid x).
\label{eq:selection-margin}
\end{equation}
Starting from the GRPO checkpoint, we apply direct preference
optimization (DPO) \citep{dpo} with the frozen GRPO policy as the
reference $\pi_{\mathrm{ref}}$:
\begin{equation}
\mathcal{L}_{\mathrm{DPO}}
=
-\mathbb{E}_{\mathcal{D}_{\mathrm{pref}}}
\log \sigma\!\left(
\beta[s_\theta(x)-s_{\mathrm{ref}}(x)]
\right).
\label{eq:dpo}
\end{equation}
The SQL candidates and execution summaries form the input.
The preference objective applies to the A/B responses.

\paragraph{One checkpoint for both roles.}
After preference training, we use the updated checkpoint
under the generation prompt to produce 16 SQL candidates.
We then use the same checkpoint under the selection prompt
to compare candidates and choose the final query.

\subsection{From Selection to Generation}
\label{sec:selection-transfer}

Execution-based RL evaluates a generated trajectory through
its final outcome. Selection training presents alternative
queries together, allowing the model to compare their
schema choices, filters, joins, and aggregations.
This comparison provides a complementary learning signal
when further execution-based training yields diminishing
returns.

Shared parameters connect the selection objective to
generation. For a single preference example, a gradient
descent step on Equation~\ref{eq:dpo} gives
\begin{equation}
\Delta\theta
=
\eta w_\theta(x)\nabla_\theta s_\theta(x),
\label{eq:preference-update}
\end{equation}
where $\eta$ is the learning rate and
\begin{equation}
w_\theta(x)
=
\beta\,\sigma\!\left(
-\beta[s_\theta(x)-s_{\mathrm{ref}}(x)]
\right)>0.
\end{equation}

Let $P_\theta(q)$ denote the probability that the tool-using
policy generates an execution-correct SQL query for question
$q$. A first-order expansion yields
\begin{equation}
\begin{aligned}
\Delta P_\theta(q)
={}&
\eta w_\theta(x)
\left\langle
\nabla_\theta P_\theta(q),
\nabla_\theta s_\theta(x)
\right\rangle \\
&+ O(\eta^2).
\end{aligned}
\label{eq:generation-transfer}
\end{equation}
Thus, selection training improves generation when its update
direction aligns with the direction that increases the
probability of correct queries. This local analysis describes
how a preference over answer labels can affect SQL generation
through the shared model parameters.

We measure this effect by comparing generation-only pass@16
before and after preference training, using the same tools
and sampling settings. For independent samples, the probability
of obtaining at least one correct query is
\begin{equation}
\operatorname{pass@16}_\theta(q)
=
1-\bigl(1-P_\theta(q)\bigr)^{16}.
\label{eq:pass16}
\end{equation}
Improved generation therefore increases the coverage of
correct candidates and raises the ceiling for final selection.

\section{Experiments}
\label{sec:experiments}

We evaluate both the final SQL predictions and the candidate
pools produced by our model. The first evaluation measures
end-to-end execution accuracy. The second examines whether
selection training improves SQL generation.

\subsection{Experimental Setup}
\label{sec:experimental-setup}

\paragraph{Dataset and metrics.}
We train on the BIRD training split and evaluate on its
development split \citep{bird}.
Execution accuracy (EX) measures whether the final SQL query
returns the same result as the reference query.
Generation-only pass@16 measures the fraction of questions
for which at least one of 16 sampled queries is
execution-correct. It also gives the oracle EX achievable
by selecting from that candidate pool.

\paragraph{Training and inference.}
We use Qwen3.8-27B as the backbone.
GRPO uses 16 rollouts per group and a batch size of 12.
Performance plateaus around step 30, and we use this checkpoint
to initialize A/B preference training.
At inference, the GRPO model achieves 73.33\% EX with
temperature zero. Our final system uses the DPO checkpoint
to generate 16 candidates and then select the final query.

\paragraph{Baselines.}
We include publicly reported results for Qwen3.8-27B from
the Artessay benchmark repository, covering its Base,
MCSSelector, and SelfDebugging configurations
\citep{artessay,mcs-sql,self-debugging}.
We also include Blender-SQL and EGV-SQL from the official
BIRD leaderboard \citep{leaderboard,egv-sql}.
Our internal comparisons evaluate three stages:
the base model with our tools, the GRPO checkpoint,
and the checkpoint after A/B selection DPO.

\subsection{End-to-End Execution Accuracy}
\label{sec:execution-results}

Table~\ref{tab:main-ex} reports execution accuracy.
The tool-using base model achieves 70.14\% EX.
GRPO improves this result to 73.33\%, a gain of 3.19
percentage points. The final system reaches 74.67\%
by generating and selecting candidates with the same
DPO-trained checkpoint.

The final system combines changes in candidate generation
and learned selection. We examine the generation component
directly in the following experiment.

\begin{table*}[t]
\centering
\begin{tabular*}{\textwidth}{@{\extracolsep{\fill}}llrr@{}}
\hline
\textbf{Method} & \textbf{Backbone}
& \textbf{Dev EX} & \textbf{Test EX} \\
\hline
\multicolumn{4}{@{}l}{\textit{Public baselines}} \\
Artessay Base
& Qwen3.8-27B & 64.34 & -- \\
Artessay MCSSelector
& Qwen3.8-27B & 64.67 & -- \\
Artessay SelfDebugging
& Qwen3.8-27B & 67.21 & -- \\
Blender-SQL
& Qwen3.8-27B + 4B & -- & 74.51 \\
EGV-SQL
& Qwen3.6-27B & 71.64 & 72.16 \\
\hline
\multicolumn{4}{@{}l}{\textit{Our system}} \\
Base + tools
& Qwen3.8-27B & 70.14 & -- \\
\quad + GRPO
& Qwen3.8-27B & 73.33 & -- \\
\quad + DPO-trained selector
& Qwen3.8-27B & 74.67 & -- \\
\hline
\end{tabular*}
\caption{Execution accuracy (\%) on BIRD.
Public baseline scores are taken from the Artessay
repository and the official BIRD leaderboard.
The GRPO result uses temperature zero.
Our final system generates 16 candidates and selects
a query using the same DPO-trained checkpoint.}
\label{tab:main-ex}
\end{table*}

\subsection{Selection Training Improves Generation}
\label{sec:generation-results}

We return the DPO checkpoint to the SQL generation task
and evaluate its candidates before selection.
Table~\ref{tab:generation-pass16} reports the results.

GRPO increases pass@16 from 80.77\% to 81.79\%.
A/B selection DPO further raises it to 82.83\%,
an additional gain of 1.04 percentage points.
The total improvement over the tool-using base model
is 2.06 percentage points.

These results show that selection training changes the
candidate pool itself. Although the preference objective
supervises A/B responses, the updated model generates
correct SQL for more questions. Teaching the model
to choose better queries therefore also improves its
generation ability and raises the ceiling for final
selection.

\begin{table}[t]
\centering
\begin{tabular}{@{}lr@{}}
\hline
\textbf{Training stage} & \textbf{Dev pass@16} \\
\hline
Base + tools              & 80.77 \\
\quad + GRPO              & 81.79 \\
\quad + A/B selection DPO & 82.83 \\
\hline
\end{tabular}
\caption{Generation-only pass@16 (\%) on BIRD Dev.
A question counts as correct when at least one of
16 sampled queries matches the reference execution
result. Scores are computed before candidate selection.}
\label{tab:generation-pass16}
\end{table}
\section{Analysis}
\label{sec:analysis}

\paragraph{Learning to choose improves what the model generates.}
A/B preference training increases generation-only pass@16
from 81.79\% to 82.83\%
(Table~\ref{tab:generation-pass16}).
The improvement appears before candidate selection:
the updated model produces correct candidates for more
questions. Selection training therefore benefits both
roles of the shared checkpoint. It improves the candidate
pool available to the selector, raising the ceiling for
the final prediction.

\paragraph{A complementary signal after the GRPO plateau.}
The GRPO run reaches a plateau around step 30.
Subsequent A/B preference training yields a further
1.04 percentage-point gain in pass@16.
One explanation is that candidate comparison provides
a complementary learning signal.
Execution rewards evaluate complete generation trajectories,
while the selection prompt presents two alternative queries
together with their execution summaries.
This context allows the model to compare concrete decisions
about joins, filters, and aggregation.
Learning these distinctions can also influence the decisions
it makes during generation, consistent with the shared-parameter
analysis in Section~\ref{sec:selection-transfer}.

\paragraph{The gap between generation and selection.}
After preference training, the model achieves 82.83\%
generation-only pass@16 and 74.67\% final EX.
The 8.16 percentage-point gap shows that correct candidates
remain unselected on some questions.
These two metrics capture the two roles of our model:
pass@16 measures the availability of correct queries,
and final EX measures the system's success in returning one.
Our results show that training selection can improve
candidate availability while also supporting final
query selection with the same checkpoint.
\bibliography{custom}

\appendix

\section{Example Appendix}
\label{sec:appendix}
\section*{Open-Source Usage Statement}

This work uses Qwen3.8-27B as its base model, released
under the Apache License 2.0. We use the model in
accordance with its license requirements. The original
project's copyright notices and license text are available
in its official repository at
\url{https://huggingface.co/Qwen/Qwen3.8-27B}.

\end{document}